%% file: main.tex
\documentclass[10pt,twocolumn,letterpaper]{article}

\usepackage{cvpr}              

\usepackage{graphicx}
\usepackage{amsmath}
\usepackage{amssymb}
\usepackage{booktabs}
\usepackage{xcolor,colortbl}
\usepackage[accsupp]{axessibility}

\definecolor{mygray}{gray}{0.88}

\input{macros}

\usepackage[pagebackref,breaklinks,colorlinks]{hyperref}

\usepackage[capitalize]{cleveref}
\crefname{section}{Sec.}{Secs.}
\Crefname{section}{Section}{Sections}
\Crefname{table}{Table}{Tables}
\crefname{table}{Tab.}{Tabs.}

\def\cvprPaperID{1533 } 
\def\confName{CVPR}
\def\confYear{2022}

\begin{document}

\title{Learning Program Representations for Food Images and Cooking Recipes}

\author{Dim~P.~Papadopoulos$^{1,3}$ \quad 
Enrique Mora$^2$ \quad 
Nadiia Chepurko$^1$ \quad 
Kuan Wei Huang$^1$ \quad \\
Ferda Ofli$^4$ \quad 
Antonio Torralba$^1$ \\ \\
$^1$ MIT CSAIL \quad
$^2$ Nestle \quad 
$^3$ DTU Compute \quad
$^4$ Qatar Computing Research Institute, HBKU
\\
{\tt\small dimp@dtu.dk,
enrique.mora@es.nestle.com,
\{nadiia,kwhuang,torralba\}@mit.edu,
fofli@hbku.edu.qa}
}

\maketitle

\input{sec0_abstract}
\input{sec1_intro}

\input{sec2_relwork}

\input{sec3_dataset}

\input{sec4_method}

\input{sec5_experiments}

\input{sec6_conc}

{\small
\bibliographystyle{ieee_fullname}
\bibliography{shortstrings.bib,dimBibTex.bib}
}

\end{document}

%% file: macros.tex
\iftrue     
\newcommand{\dimpp}[1]{\textcolor{blue}{[DP: #1]}}
\newcommand{\antonio}[1]{\textcolor{orange}{[AT: #1]}}
\newcommand{\enrique}[1]{\textcolor{green}{[EM: #1]}}
\newcommand{\nadiia}[1]{\textcolor{magenta}{[NC: #1]}}
\newcommand{\ferda}[1]{\textcolor{cyan}{[FO: #1]}}
\newcommand{\status}[1]{\textcolor{orange}{[STATUS: #1]}}
\else
\newcommand{\dimpp}[1]{\textcolor{blue}{\noindent}}
\newcommand{\antonio}[1]{\textcolor{orange}{\noindent}}
\newcommand{\enrique}[1]{\textcolor{green}{\noindent}}
\newcommand{\nadiia}[1]{\textcolor{magenta}{\noindent}}
\newcommand{\ferda}[1]{\textcolor{cyan}{\noindent}}
\newcommand{\status}[1]{\textcolor{orange}{\noindent}}
\fi

\newcommand{\mypar}[1]{\noindent\vspace{-0mm}\textbf{#1}}

%% file: sec0_abstract.tex
\begin{abstract}

 In this paper, we are interested in modeling a how-to instructional procedure, such as a cooking recipe, with a meaningful and rich high-level representation. Specifically, we propose to represent cooking recipes and food images as cooking programs. Programs provide a structured representation of the task, capturing cooking semantics and sequential relationships of actions in the form of a graph. This allows them to be easily manipulated by users and executed by agents. To this end, we build a model that is trained to learn a joint embedding between recipes and food images via self-supervision and jointly generate a program from this embedding as a sequence. To validate our idea, we crowdsource programs for cooking recipes and show that: (a) projecting the image-recipe embeddings into programs leads to better cross-modal retrieval results; (b) generating programs from images leads to better recognition results compared to predicting raw cooking instructions; and (c) we can generate food images by manipulating programs via optimizing the latent code of a GAN.
 Code, data, and models are available online\footnote{\url{http://cookingprograms.csail.mit.edu}}.

\end{abstract}

%% file: sec1_intro.tex
\section{Introduction}
\label{sec:intro}


Food is an important part of our lives. Imagine an AI agent that can look at a dish and recognize ingredients and reliably reconstruct the exact recipe of the dish, or another agent that can read, interpret and execute a cooking recipe to produce our favorite meal.
%
Computer vision community has long studied image-level food classification~\cite{bossard14eccv,chen2017chinesefoodnet,kaur2019foodx,lee2018cleannet,liu2016deepfood,min2020isia}, and only recently focused on understanding the mapping between recipes and images using multi-modal representations~\cite{fu2020mcen,marin2019recipe1m+,salvador21cvpr,salvador17cvpr,zhu19cvpr}. However, retrieval systems are limited to the existing database and usually fail for queries outside of this database while generating the full recipe from an image remains a challenge~\cite{salvador19cvpr}.

\begin{figure}[t]
\center
\includegraphics[width=\linewidth]{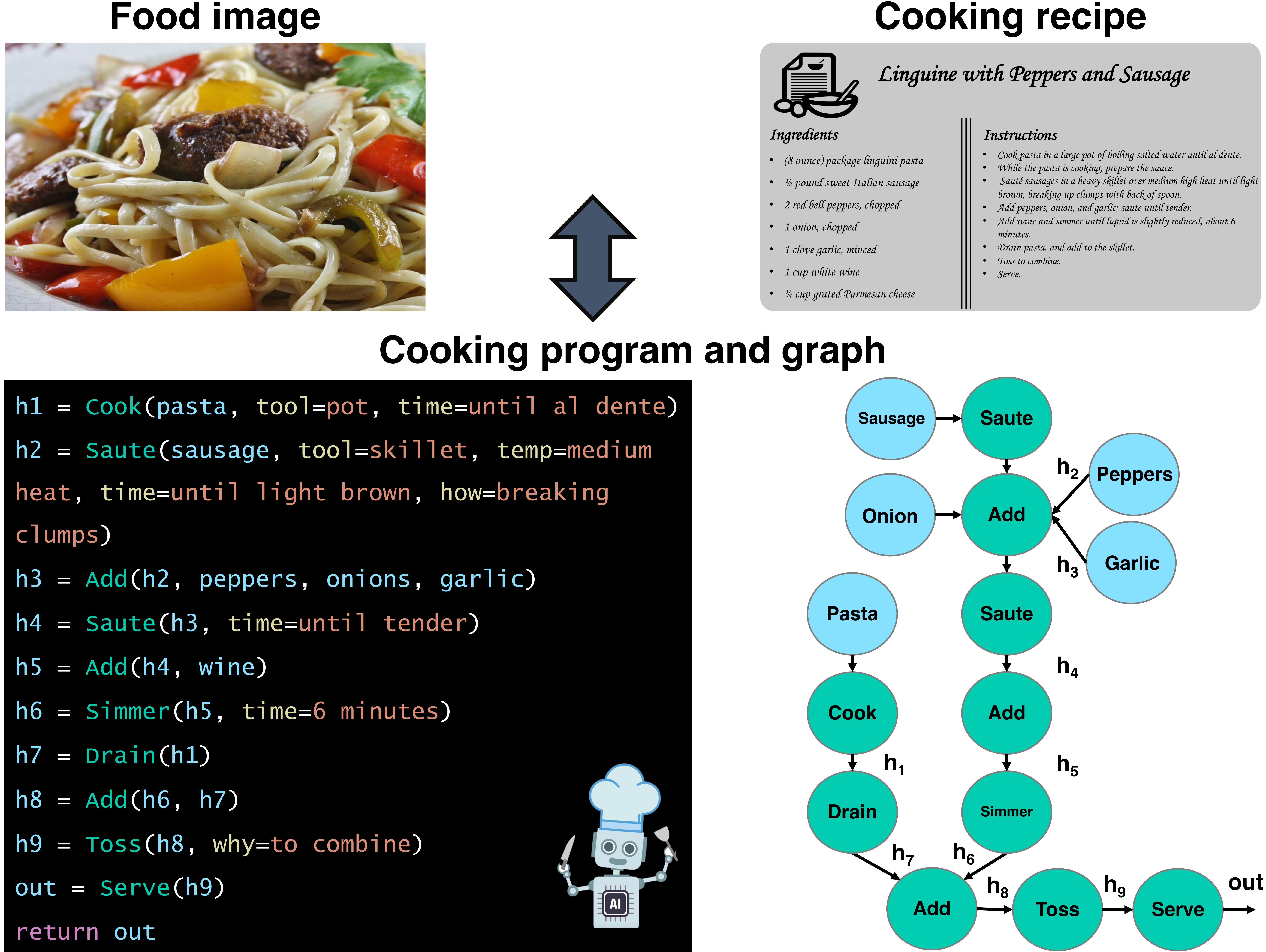}
\caption{\small \textbf{Cooking programs.} We learn cooking programs from food images and recipes. Programs provide a structured representation of the cooking procedures which can also be represented as graphs (for brevity, we only show action and ingredient nodes).}
\label{fig:teaser}
\end{figure}

\begin{figure*}[t]
\center
\includegraphics[width=\linewidth]{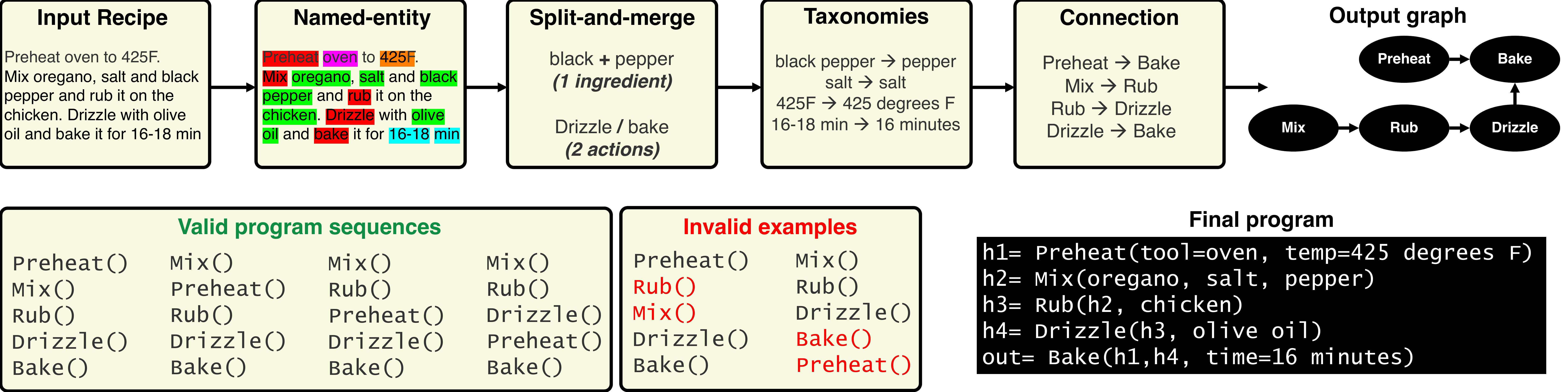}
\caption{\small \textbf{Annotation of cooking programs.} (Top) We obtain a graph from an input recipe via named-entity, split-and-merge parsing, taxonomy dictionaries and connection annotation. (Bottom) We obtain all valid program sequences from the graph and the final program.}
\label{fig:programAnnotation}
\end{figure*}

Cooking recipes are step-by-step instructional procedures which we propose to represent as programs, capturing all the cooking semantics and relationships. 
A program contains a sequence of actions that can be written as functions (e.g., \texttt{Cook()}, \texttt{Add()}). Each action operates on specific ingredients under certain conditions, such as time or tool (e.g., \texttt{Cook(pasta, time=`10 minutes', tool=`pot')}). A program also captures the sequential dependency of the actions by maintaining their input-output connections. Note that even though cooking actions are often performed sequentially in time, their connections are not necessarily sequential. 
The program can also be represented as a graph where each function and parameter is a node, while the edges are the function connections or the connections between the parameters and the actions (Fig.~\ref{fig:teaser}).

Our goal is to generate cooking programs conditioned on food images or cooking recipes. We build a model that leverages the natural pairing of food images and recipes by learning a joint embedding 
using a vision and a text encoder. The visual and text embeddings are then used in a program decoder to generate cooking programs. Our model is trained end-to-end by jointly optimizing a ranking loss between the visual and text representations and two losses on the program sequence predictions. Because the sequence of some actions can be permuted without violating the input-output connections between the functions, we generate the set of all valid program sequences for each recipe (Fig.~\ref{fig:programAnnotation}) and design a loss that operates on this set. At test time, the model can not only perform image-to-recipe retrieval tasks but can also predict the cooking program from an image or a recipe.


To validate our idea, we first crowdsource programs for cooking recipes selected from the Recipe1M dataset~\cite{salvador17cvpr} using carefully designed tasks that can be easily performed by naive annotators.
%
%
Experimental results show that our model leads to better cross-modal retrieval when it is jointly trained to generate programs. Moreover, generating programs leads to better food recognition results compared to predicting the raw cooking instructions. Finally, we show how to generate food images by manipulating cooking programs via optimizing the latent code of a GAN.


%% file: sec2_relwork.tex
\section{Related Work}
\label{sec:relwork}


\mypar{Food recognition.}
Since the Food-101 dataset~\cite{bossard14eccv}, food image analysis has become an established problem in computer vision~\cite{chen2017chinesefoodnet,engilberge2018finding,kaur2019foodx,lee2018cleannet,marin2019recipe1m+,min2020isia,salvador19cvpr,salvador21cvpr}. 
While early work focused mostly on image-level food categorization~\cite{bossard14eccv,chen2016deep,liu2016deepfood,min2017delicious}, recent studies performed more fine-grained analyses such as ingredient recognition~\cite{chen2016deep,salvador19cvpr}, nutrition and calorie estimation~\cite{korpusik2017spoken,meyers2015im2calories}, food logging~\cite{merler2016snap}, and image generation~\cite{han2020cookgan,papadopoulos19cvpr,zhu2020cookgan}. 
Lately, cross-modal analysis of recipes and food images has become popular, thanks to the Recipe1M~\cite{salvador17cvpr}, where the authors tackled the image-to-recipe retrieval problem~\cite{salvador17cvpr,marin2019recipe1m+}. 
Several studies improved the performance, including ACME~\cite{wang2019learning}, $R^2$-GAN~\cite{zhu19cvpr}, MCEN~\cite{fu2020mcen}, SCAN~\cite{wang2021cross}, among others~\cite{CarvalhoM:SIGIR18,chen2018deep,fain2019dividing,pham2021chef,salvador21cvpr}. 
However, we aim to go beyond this task and understand the joint latent space of recipes and images so that we can generate new recipes from images and vice versa. Along these lines, \cite{salvador19cvpr} presented a two-step approach for predicting recipes from images: they first extract a list of ingredients given an image, and then, train a decoder that generates a recipe given both the image and the list of ingredients.

\mypar{Recipe parsing.}
The task here is to parse a recipe and segment it into a sequence of individual actions. To achieve this, existing studies explored flow graphs~\cite{mori2014flow,kiddon2015mise,yamakata2020english,nishimura2020visual}, tree-based solutions~\cite{jermsurawong2015predicting,chang2018recipescape}, or extracting knowledge from cooking videos and transcripts without using the recipe text~\cite{xu2020benchmark}.
However, the datasets used in these studies are small (less than 300 recipes) and limited to only verbs and ingredients~\cite{chang2018recipescape,jermsurawong2015predicting}. More importantly, the parsing models require a specific and curated custom format for the recipes~\cite{jermsurawong2015predicting,kiddon2015mise}. Instead, we propose a program representation and describe how we can obtain programs for uncurated recipes~\cite{salvador17cvpr} using efficient and scalable interfaces.

\begin{figure}[t]
\center
\includegraphics[width=0.95\linewidth]{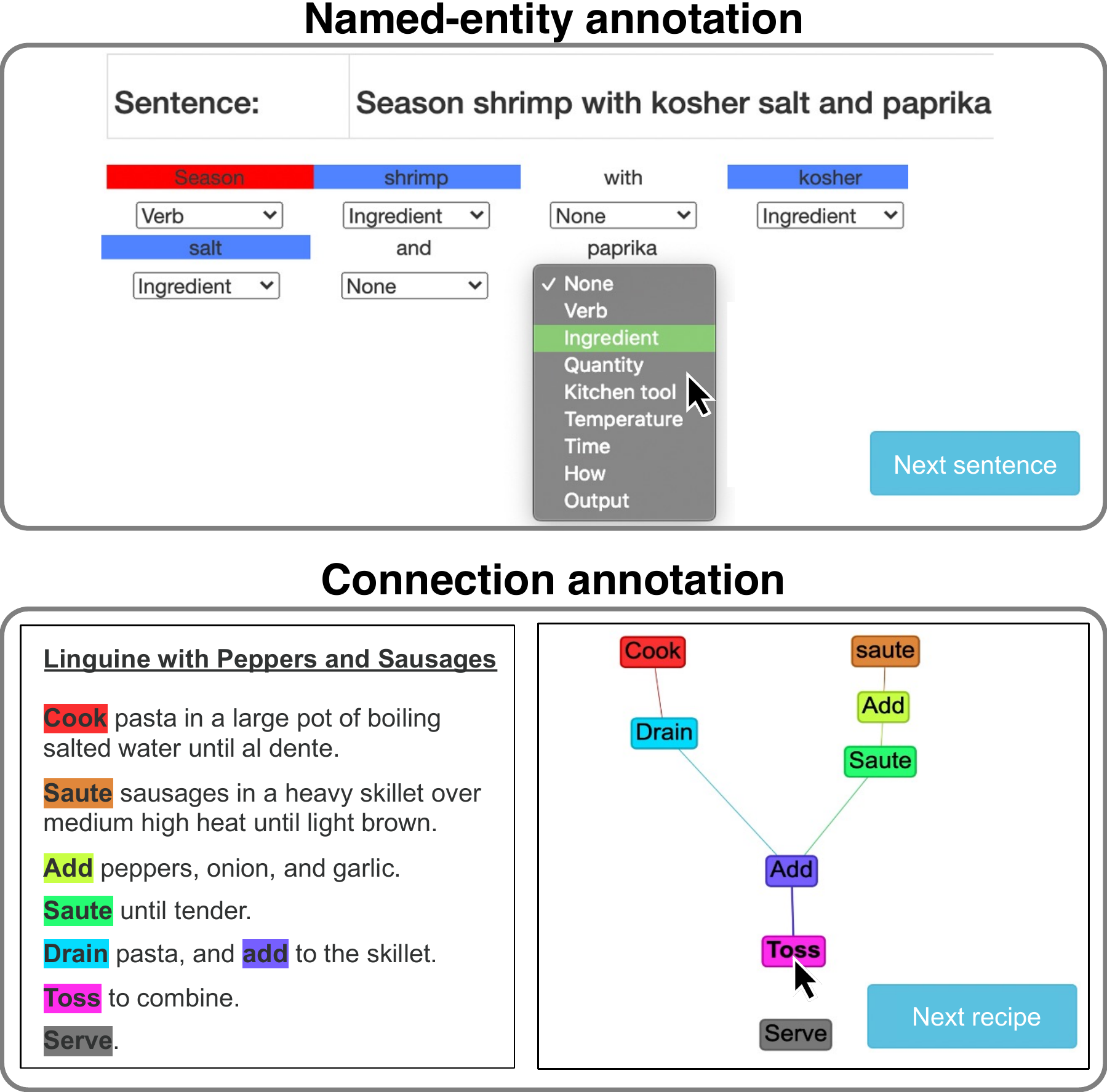}
\caption{\small \textbf{AMT interfaces.} \textbf{(Named-entity annotation)} We ask the annotators to tag every word of a cooking sentence. \textbf{(Connection annotation)} We ask the annotators to connect two actions (colored nodes) if there is an in-out relationship between them.}
\label{fig:programAMT}
\end{figure}


\mypar{Program representation.} 
Our work is also related to approaches that represent activities as programs and learn them from visual inputs (typically videos)~\cite{alayrac2016unsupervised,nyga2012everything,puig2018virtualhome,yang2014manipulation,yang2015robot}. This includes learning actions and objects by watching cooking videos~\cite{yang2014manipulation,yang2015robot} or a sequence of atomic actions from instruction videos~\cite{alayrac2016unsupervised}. In~\cite{puig2018virtualhome}, programs for household activities were used to train model that generate programs from videos or from natural language descriptions.
Program generation methods have also been proposed for various tasks, such as visual question answering ~\cite{johnson2017inferring,yi2018neural,mao2019neuro}, fact verification~\cite{chen2020tabfact}, and geometry problem solving~\cite{lu2021inter}.

%% file: sec3_dataset.tex
\section{Cooking programs}
\label{sec:dataset}


In this section, we describe our program representation and show how we design a scalable crowdsourcing protocol to efficiently annotate recipes and obtain cooking programs.

\subsection{Program scheme and graph representation}
\label{sec:dataset_program}

Cooking recipes consist of a title, a list of ingredients, and a list of instruction steps. Each step describes a cooking action that operates on ingredients using tools under certain conditions (e.g., time or temperature). This action results in an intermediate output that is used as an input in one of the following steps. This interesting structure, which resembles to the source code of a program, motivates us to use programs to represent cooking recipes (Fig.~\ref{fig:teaser}). 

A program contains a sequence of functions that correspond to cooking actions. Each function takes as input a list of input variables (ingredients) and parameters (e.g., ingredient quantities or the way the action is performed, such as using a tool). For example, the sentence
``Bake the chicken in the oven for 10 minutes at 400 degrees F''
can be written as:
%
\texttt{h = Bake(chicken, tool=oven, time=10 minutes, temp=400 degrees F);}
The output of the function is denoted as $h$.
For a full program, we also capture the input-output connections between the individual commands. Even though a recipe is performed sequentially in time, the output of an action is not always used as an input to the next action. Fig.~\ref{fig:teaser} shows an example where the recipe consists of two sub-recipes that are combined at the end (i.e, cook pasta and prepare the sauce). These connections are captured by using the latent variables $h$ as inputs to the corresponding functions (e.g. \texttt{Drain(h1)}).

\begin{figure*}[t]
\center
\includegraphics[width=\linewidth]{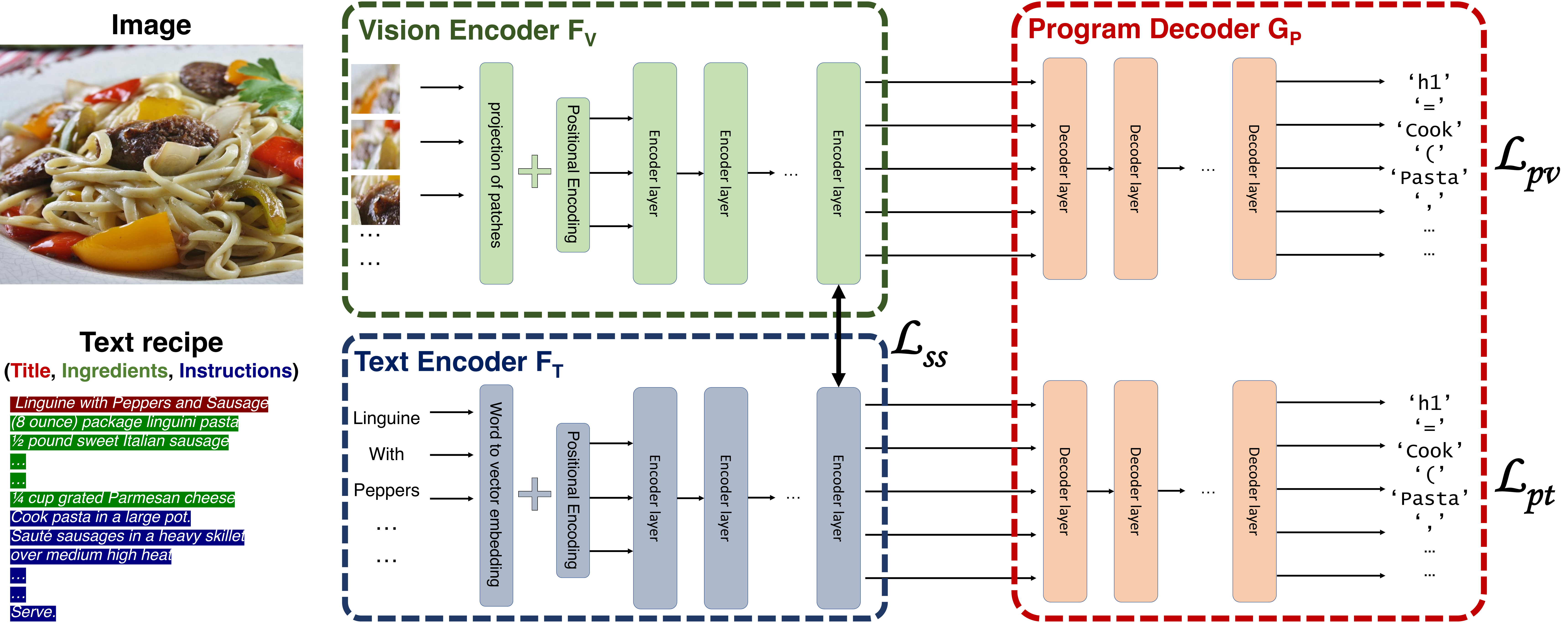}
\caption{\small \textbf{Generating cooking programs from food images or cooking recipes.} Our model is trained end-to-end to project image and recipe features from a vision and a text encoder into a common space and to jointly generate programs as a sequence of commands using a program decoder conditioned on the image and text features.}
\label{fig:main_figure}
\end{figure*}

\mypar{Cooking graph.} 
Cooking programs can be also represented as graphs $\mathcal{G=(V,E)}$ (Fig.~\ref{fig:teaser} (right)). Variables and functions are represented as vertices $\mathcal{V}$, while the edges $\mathcal{E}$ are the connections between the functions and their inputs.
The graph shows that the sequence of actions can be permuted without violating input-output connections (i.e, the edges $\mathcal{E}$). We can compute all possible permutations that lead to valid, executable programs, i.e., $h_i$ cannot be used as input before being computed. The bottom of Fig.~\ref{fig:programAnnotation} shows all valid permutations where the function \texttt{Preheat()} can be executed at any point before baking the chicken. In Sec.~\ref{sec:method} we use this set of permutations to train our proposed model.

\mypar{Program taxonomy.} Parsing the recipes creates huge vocabularies for each category of the programs (Recipe1M contains 230k unique ingredients, 20k actions and 40k tools).
To create a fixed and semantically meaningful vocabulary for each category, we follow a semi-automatic procedure. We describe here the process for the ingredients and we follow a similar one for the others:
First, we use Sentence-BERT~\cite{reimers19sentence} and extract features for each unique ingredient. Then, we cluster them with K-Means using a high number of clusters (2,000) so that each cluster contains only semantically identical ingredients. Finally, for each cluster, we manually check its nearest neighbor and merge them if they are semantically almost identical. We repeat this iteratively until clusters cannot be merged anymore. Overall, this leads to 
514 \textit{ingredient}, 
60 \textit{action}, 
55 \textit{tool}, 
130 \textit{quantity}, 
60 \textit{temperature}, 
152 \textit{time}, 
105 \textit{how},
112 \textit{why} and 
220 \textit{output} clusters.

\subsection{Program annotation}
\label{sec:dataset_anno}

Crowdsourcing cooking programs is challenging as most naive annotators have no programming experience. For this reason, we split the process into four simple steps: (a) named-entity annotation, (b) split-and-merge parsing, (c) connection annotation and (d) program taxonomy (Fig.~\ref{fig:programAnnotation}).

\mypar{Named-entity annotation.}
In this step, we provide annotators a cooking sentence, and ask them to classify (tag) every word as one of the following categories: \textit{cooking action}, \textit{ingredient}, \textit{quantity}, \textit{kitchen tool}, \textit{temperature}, \textit{time}, \textit{how}, \textit{why}, and \textit{output} (Fig.~\ref{fig:programAMT} (left)).
The annotators first read instructions with the definition of each category and several examples. To ensure high quality responses, we design a protocol that follows common quality control mechanisms~\cite{kuznetsova18arxiv,papadopoulos17cvpr,russakovsky15ijcv,sorokin08cvprw,vondrick13ijcv,zhou17pami} and monitor the performance of the annotators by using hidden pre-tagged sentences.

\mypar{Split and merge.}
In this step, we perform two parsing operations without any human intervention. First, we split sentences that contain more than one action (e.g., ``drizzle with olive oil and bake it for 16-18 minutes'' in Fig.~\ref{fig:programAnnotation}) so that every sentence contains only one action. Next, we merge tagged words into entities. For example, the phrase ``mix fresh oregano, salt and black pepper" has five ingredient words but there are only three ingredients. Black pepper and fresh oregano are merged into the final variables.

\mypar{Connection annotation.}
In this step, we provide annotators a recipe with highlighted actions and a panel with these actions as nodes (Fig.~\ref{fig:programAMT} (right)), and ask them to connect two nodes with an edge if there is an input-output relationship between them. 
To ensure high quality, we provide instructions and follow standard quality control mechanisms.

\mypar{Program taxonomies.}
In this step, we follow the variable taxonomies (Sec.~\ref{sec:dataset_program}) and for each category (i.e, actions, ingredients, tools, etc.) we map all obtained values of each node into the corresponding categories from our fixed program vocabulary (e.g., ``black pepper''$\rightarrow$``pepper'' in Fig.~\ref{fig:programAnnotation}).

\mypar{Final program.}
Finally, we follow the rules described in Sec.~\ref{sec:dataset_program} to create the final program and the corresponding graph. The output of the last action is denoted as \texttt{out}.

\begin{table*}[t]
\centering
\resizebox{0.89\textwidth}{!}{
\begin{tabular}{c c c c c c c c c c c}
\bottomrule
\rowcolor{mygray}
\textbf{Recipe} & \textbf{Program} & \textbf{Encoder}  &\multicolumn{4}{c}{\textbf{image-to-recipe}} & \multicolumn{4}{c}{\textbf{recipe-to-image}} \\
\rowcolor{mygray}
\textbf{Component} & \textbf{Loss} & \textbf{Layers} & \textbf{medR} & \textbf{R@1}& \textbf{R@5} & \textbf{R@10} & \textbf{medR} & \textbf{R@1} & \textbf{R@5} & \textbf{R@10} \\
\toprule
Title &  & 8 (small) & 4.6 & 24.0 & 54.4 & 67.8 & 4.5 & 23.8 & 54.2 & 67.5 \\
Ingredients &  & 8 (small) & 2.8 & 39.1 & 68.1 & 80.3 & 3.0 & 38.9 & 68.4 & 80.5 \\
Instructions &  & 8 (small) & 3.1 & 36.6 & 65.2 & 76.8 & 3.0 & 36.5 & 65.8 & 76.9  \\
Title+Ingr &  & 8 (small) & 2.0 & 43.6 & 75.6 & 85.0 & 2.0 & 44.9 & 76.2 & 85.4  \\
Title+Ingr+Inst &  & 8 (small) & 1.0 & 53.5 & 81.8 & 89.2 & 1.0 & 53.1 & 82.0 & 89.6 \\ 
\midrule
Title+Ingr+Inst & \checkmark  & 8 (small) & 1.0 & 58.6 & 85.7 & 91.7 & 1.0 & 58.2 & 85.5 & 92.0 \\ 
Title+Ingr+Inst & \checkmark  & 12 (base) & \textbf{1.0} & \textbf{66.9} & \textbf{90.9}& \textbf{95.1} & \textbf{1.0} & \textbf{66.8} & \textbf{89.8} & \textbf{94.6}  \\
\bottomrule \\
\bottomrule
\rowcolor{mygray}\multicolumn{3}{l}{\textbf{State-of-the-art}} &\multicolumn{4}{c}{\textbf{image-to-recipe}} & \multicolumn{4}{c}{\textbf{recipe-to-image}} \\
\rowcolor{mygray}\multicolumn{2}{l}{\textbf{cross-modal retrieval}}& \textbf{Img Encoder} & \textbf{medR} & \textbf{R@1}& \textbf{R@5} & \textbf{R@10} & \textbf{medR} & \textbf{R@1} & \textbf{R@5} & \textbf{R@10} \\
\toprule
\multicolumn{2}{l}{Salvador CVPR 17~\cite{salvador17cvpr}}
& ResNet-50 & 5.2 & 24.0 & 51.0 & 65.0 & 5.1 & 25.0 & 52.0 & 65.0 \\
\multicolumn{2}{l}{Chen ACM MM 18~\cite{chen2018deep}} & ResNet-50 & 4.6 & 25.6 & 53.7 & 66.9 & 4.6 & 25.7 & 53.9 & 67.1\\
\multicolumn{2}{l}{Carvalho SIGIR 18~\cite{CarvalhoM:SIGIR18}}& ResNet-50  & 2.0 & 39.8 & 69.0 & 77.4 & 1.0 & 40.2 & 68.1 & 78.7 \\
\multicolumn{2}{l}{Zhu CVPR 19~\cite{zhu19cvpr}}& ResNet-50  & 2.0 & 39.1 & 71.0 & 81.7 & 2.0 & 40.6 & 72.6 & 83.3 \\
\multicolumn{2}{l}{Fu CVPR 20~\cite{fu2020mcen}}& ResNet-50  & 2.0 & 48.2 & 75.8 & 83.6 & 1.9 & 48.4 & 76.1 & 83.7 \\
\multicolumn{2}{l}{Pham AAAI 21~\cite{pham2021chef}}& ResNet-50  & 1.6 & 49.7 & 79.3 & 86.3 & 1.6 & 50.1 & 79.0 & 86.4 \\ 
\multicolumn{2}{l}{Wang CVPR 19~\cite{wang2019learning}}& ResNet-50  & 1.0 & 51.8 & 80.2 & 87.5 & 1.0 & 52.8 & 80.2 & 87.6\\
\multicolumn{2}{l}{Wang TMM 21~\cite{wang2021cross}}& ResNet-50 & 1.0 & 54.0 & 81.7 & 88.8 & 1.0 & 54.9 & 81.9 & 89.0\\
\multicolumn{2}{l}{Fain arXiv 19~\cite{fain2019dividing}}& ResNeXt-101  & 1.0 & 60.2 & 84.0 & 89.7 & -- & -- & -- & --\\
\multicolumn{2}{l}{Salvador CVPR 21~\cite{salvador21cvpr}}& ResNet-50  & 1.0 & 60.0 & 87.6 & 92.9 & 1.0 & 60.3 & 87.6 & 93.2 \\
\multicolumn{2}{l}{Salvador CVPR 21~\cite{salvador21cvpr}}& ViT-B/16  & 1.0 & 63.2 & 88.3 & 93.1 & -- & -- & -- & -- \\
\midrule
\multicolumn{2}{l}{\textbf{Ours}} & \textbf{ViT-B/16} & \textbf{1.0} & \textbf{66.9} & \textbf{90.9}& \textbf{95.1} & \textbf{1.0} & \textbf{66.8} & \textbf{89.8} & \textbf{94.6}  \\
\bottomrule 
\end{tabular}
}
\caption{\small \textbf{Cross-modal retrieval results on Recipe1M.} At the top part of the table, we report ablation studies of our model, while at the bottom part, we compare our results with the state-of-the-art cross-modal retrieval approaches.}
\label{table:retrieval}
\end{table*}

\subsection{Program collection}
\label{sec:dataset_stats}

We ran experiments on Amazon Mechanical Turk (AMT) and collected programs for 3,708 recipes selected from the Recipe1M dataset~\cite{salvador17cvpr}\footnote{Human subject experiments were conducted with an IRB approval.}. This translates to 42,473 sentences with 478,285 tagged words and 54,154 annotated edge connections. An annotated example is shown in Fig.~\ref{fig:programAnnotation}. 

\mypar{Annotation time and quality.}
The median response time of the named-entity task was 17 s per sentence, while the time of the connection task was 75 s per recipe. The annotated recipes have on average 11 sentences leading to a total annotation time of 4 minutes per recipe.
%
The total cost for annotating our programs was about \$2,000 (hourly wage about \$8).
%
To analyze and quantify the quality of our annotations, we annotate 50 recipes (550 sentences) once more using different annotators and measure the human agreement. The human agreement of the named-entity task is 97.9\% (i.e., words with the same annotation label) . For the connection task, we found that 7.3\% of the connections do not appear in both annotations (agreement of 92.7\%).

%% file: sec4_method.tex
\section{From food images and recipes to programs}
\label{sec:method}


We introduce the novel task to generate cooking programs from food images or cooking recipes. In Sec.~\ref{sec:method_overview}, we present the model architecture, while in Sec~\ref{sec:method_learning}, we explain the training process and the objective functions.

\subsection{Model architecture}
\label{sec:method_overview}

Given a set of images $\mathcal{I}$ paired with their recipes $\mathcal{R}$ and a set of programs ${P}$, our goal is to learn how to generate a cooking program conditioned on an image or a recipe. Our model consists of three components: (a) a vision encoder, (b) a text encoder, and (c) a program decoder (Fig.~\ref{fig:main_figure}). The model 
is trained to embed the images and recipes into the same space via self-supervision and to generate a program from an image or a recipe embedding. 

\mypar{Vision encoder.}
Food images are fed into the vision encoder ${F}_{V}$ based on the Vision Transformer (ViT)~\cite{dosovitskiy2020image}.
The image is split into fixed-size patches (tokens). After a linear projection and adding position embedding, these tokens are fed to a stack of $k_v$ Transformer encoder layers~\cite{vaswani17nips}. Unlike the original ViT where the image representation is obtained from the ``classification token", we obtain it from the features of all patches after average pooling.

\mypar{Text encoder.}
\label{sec:method_text}
The text from cooking recipes is fed into the text encoder ${F}_{T}$~\cite{vaswani17nips}. The architecture of ${F}_{T}$ is similar to ${F}_{V}$ with the difference that here the words play the role of the tokens fed into $k_t$ identical Transformer layers~\cite{vaswani17nips}. Once more, the final recipe representation is obtained after an average pooling layer on top of the token embeddings.


\mypar{Program decoder.}
The program decoder ${G}_{P}$ consists of $k_p$ transformer decoder layers~\cite{vaswani17nips}.
The features obtained from ${F}_{V}$ are fed into the multi-head attention of each layer following the standard attention mechanism~\cite{vaswani17nips}. This results in a predicted program sequence ${P}_{v}$ given the image.
We repeat the approach for the text part. As such, the features obtained from ${F}_{T}$ are fed to ${G}_{P}$ to predict a program sequence ${P}_{t}$. 
During inference, a program can be predicted given either a food image or a recipe using the corresponding features and the same decoder ${G}_{P}$ (shared weights).

\subsection{Training model and loss functions}
\label{sec:method_learning}

We train our model end-to-end to jointly learn to project the image features from 
${F}_{V}$ and the text features from 
${F}_{T}$ into a common space through a self-supervised loss $\mathcal{L}_{ss}$ and to generate programs matching the ground-truth ones from the food images $\mathcal{L}_{pv}$ and the cooking recipes $\mathcal{L}_{pt}$.




\begin{table}[t]
\centering
\resizebox{0.9\linewidth}{!}{
\begin{tabular}{c c c c c c}
\bottomrule
\rowcolor{mygray}
\textbf{Loss} & \textbf{ViT}  &\multicolumn{3}{c}{\textbf{image-to-recipe}} \\
\rowcolor{mygray}
\textbf{Hyperparameters} & \textbf{Features}& \textbf{R@1} & \textbf{R@5} & \textbf{R@10}
\\ \toprule
$\lambda_{pv}=0.1$, $\lambda_{pt}=0.1$ & average &\textbf{58.6} & \textbf{85.7} & \textbf{91.7}\\ 
\midrule
$\lambda_{pv}=0.1$, $\lambda_{pt}=0.1$ & cls token & 56.1 & 83.8 & 90.1\\ 
\midrule
$\lambda_{pv}=0$, $\lambda_{pt}=0$ & average & 53.5 & 81.8 & 89.2\\ 
$\lambda_{pv}=1$, $\lambda_{pt}=1$ & average & 58.0 & 85.1 & 90.9 \\ 
$\lambda_{pv}=10$, $\lambda_{pt}=10$ & average & 56.4 & 83.7 & 89.6 \\ 
\bottomrule
\end{tabular}
}
\caption{\small \textbf{Ablation study} on ViT feature representation and loss hyperparameters. All models use encoders with 8 layers (small).}
\label{table:ablation}
\end{table}

\begin{figure*}[t]
\center
\includegraphics[width=\linewidth]{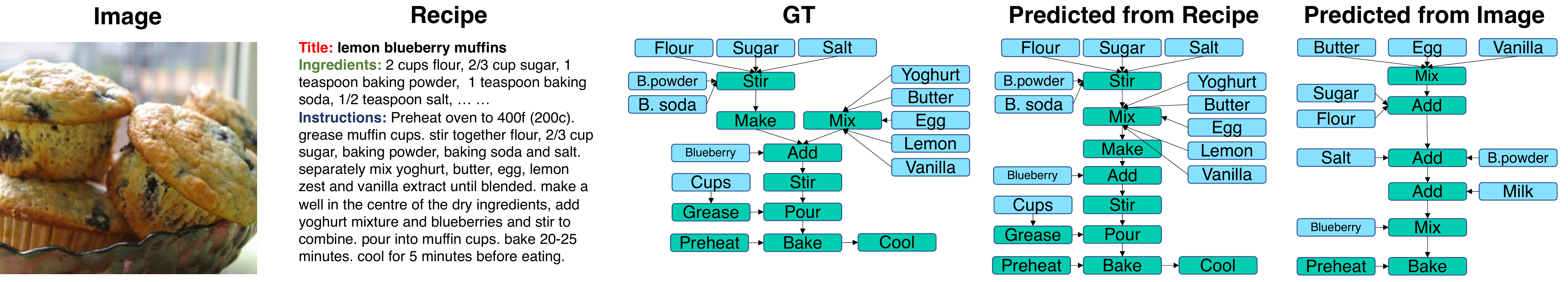}
\caption{\small \textbf{Generating programs from food images and recipes.} We show the output graphs of the two generated programs conditioned on the input recipe or the input image. For visualization purposes, we only show action (green) and ingredient (cyan) nodes.}
\label{fig:graph_example}
\end{figure*}

\mypar{Self-supervised triplet loss.}
We follow prior work and use a bi-directional max-margin triplet ranking loss to project images and text descriptions into a joint space~\cite{karpathy2015deep,karpathy2014deep,socher2014grounded,wang2016learning} due to its recent success at the image-to-recipe retrieval task~\cite{CarvalhoM:SIGIR18,salvador21cvpr,wang2019learning,zhu19cvpr}.
The loss is based on the triplet ranking loss~\cite{schroff15facenet,schultz04learning} $\mathcal{L}_{t}(a, p, n) = max(0, s(f(a), f(n)) - s(f(a), f(p)) + m)$ 
where $a$ is an anchor input, $p$ and $n$ are positive and negative samples, $s$ is a similarity function, $f$ is an embedding and $m$ is a fixed constant margin. 

For a mini-batch with size $N$ of image-recipe pairs $\{I_i,R_i\}_{i=1}^N$, we obtain the feature representations of the image $I_i$ and a recipe $R_j$ from the visual $F_V$ and text encoder $F_T$ as $F_V(I_i)$ and $F_T(R_j)$, respectively.
The image-recipe pair is considered as positive when $i=j$ and as negative otherwise. 
The final self-supervised bi-directional loss of the mini-batch, which computes $\mathcal{L}_{t}$ twice, is given by:
\begin{align}
    \mathcal{L}_{ss} = \frac{1}{N}\sum_{i=1,j \neq i}^{N} max(0, s(i,j) - s(i,i) + m) + \nonumber  \\
    + max(0, s(j,i) - s(i,i) + m)
\end{align}
where $s(i,j)$=$s(F_V(I_i), F_T(R_j))$ denotes the cosine similarity between image $I_i$ and a recipe $R_j$.

\mypar{Program prediction loss.}
Each recipe $R_i$ is a sequence of $K$ sentences $(r^1, r^2, ..., r^K)$. The program $P_i$ is a sequence of $L$ programming function commands $(p^1, p^2, ..., p^L)$ following the ordering of the recipe instructions (Fig.~\ref{fig:teaser}). However, as described in Sec.~\ref{sec:dataset_program}, some of the commands $p^k$ can be permuted without violating the input-output connections between the functions. We denote as $\mathcal{P}_{i}=(P_i^0, P_i^1, ... P_i^\beta)$ the set of all $\beta$ valid permutations of the sequence $P_i$. 
At each time step, $G_p$ predicts a probability distribution for the output tokens, which is typically followed by a softmax and a cross-entropy loss between the predicted and the target sequence. Here, we expand the cross-entropy loss to handle a set of multiple candidate sequences $\mathcal{P}$.

Let $\Pi_i^I=G_p(F_V(I_i))$ be the generated program condition on $I_i$ and $\Pi_i^R=G_p(F_T(R_i))$ be the generated program condition on $R_i$.
The loss $\mathcal{L}_{pv}$
for a mini-batch is given by:
\begin{equation}
    \mathcal{L}_{pv} = \frac{1}{N}\sum_{i=1}^{N} 
    \min_{j \in [1,\beta]} \mathcal{L}_{ce}(\Pi_i^I,P_i^j)
\label{eq:lpv}
\end{equation}
where $\mathcal{L}_{ce}(\Pi_i^I,P_i^j)$ is the cross-entropy loss between $\Pi_i$ and the $j^{th}$ target program sequence $P_i^j$. $\mathcal{L}_{pt}$  is given by eq.~\eqref{eq:lpv} by replacing $\Pi_i^I$ with $\Pi_i^R$.

\mypar{Final loss.} The full objective function $\mathcal{L}$ of our model is defined as $\mathcal{L}$ = $\lambda_{ss}\mathcal{L}_{ss}$ + 
$\lambda_{pv}\mathcal{L}_{pv}$ + $\lambda_{pt}\mathcal{L}_{pt}$, where $\lambda_{ss}$, $\lambda_{pv}$, $\lambda_{pt}$ are hyperparameters that control the relative importance of the image-recipe loss to the program prediction losses.

\begin{table}[t]
\centering
\resizebox{1\linewidth}{!}{
\begin{tabular}{l c c c c}
\bottomrule
\rowcolor{mygray}\multicolumn{5}{c}{\textbf{Input: Food images}} \\
\toprule
&  \textbf{Ingredients} & \textbf{Actions} & \textbf{Tools} & \textbf{Full graph$^*$}  \\
&  \textbf{(F1 $\uparrow$)} & \textbf{(F1 $\uparrow$)} & \textbf{(F1 $\uparrow$)} & \textbf{(GED $\downarrow$)}  \\
\midrule
\textbf{Random image} & 12.6 & 14.6 & 14.2 & 102.1  \\
 \textbf{Retrieved image}  & 39.4 &  51.6 & 66.9 & 79.1 \\
 \textbf{NN (oracle)}  & 53.5 &  66.5 & 81.1 & 62.2 \\
 \textbf{Instructions} & 28.5 & 38.3 &  50.5 & --  \\
 \textbf{Programs (CE)}  & 52.8 & 64.5 &  78.1 & 72.1 \\
 \textbf{Programs (minCE)} & 53.5 & 64.7 &  78.1 & 67.2 \\
\bottomrule \\ 
\bottomrule 
\rowcolor{mygray}\multicolumn{5}{c}{\textbf{Input: Cooking recipes}} \\
\toprule
& \textbf{Ingredients} & \textbf{Actions} & \textbf{Tools} & \textbf{Full graph$^*$} \\
&  \textbf{(F1 $\uparrow$)} & \textbf{(F1 $\uparrow$)} & \textbf{(F1 $\uparrow$)} & \textbf{(GED $\downarrow$)}\\
\midrule
\textbf{Random recipe} & 12.4 &  14.5 & 14.2 & 101.5 \\
\textbf{Retrieved recipe} & 43.4 &  55.2 & 74.2 & 67.1  \\
\textbf{NN (oracle)} & 53.5 &  66.5 & 81.1 & 57.2 \\
\textbf{Instructions}& 41.6 & 49.3 & 66.6 & --  \\
\textbf{Programs (CE)} & 75.4 & 83.1 &  83.8 & 19.1 \\
\textbf{Programs (minCE)} & 75.5 & 83.1 &  84.1 & 16.8 \\
\bottomrule 
\end{tabular}}
\caption{\small \textbf{Evaluation of the predicted programs from images (top) and from recipes (bottom)}. For the full program, we report the graph edit distance (GED) between the ground-truth and the predicted graphs. We also extract the ingredients, actions and tools from the programs and report the F1 score with respect to the ground truth. $^*$Note that due to the high computational cost of graph matching, GED is computed only on 5\% of the test set.}
\label{table:ingredients}
\end{table}



%% file: sec5_experiments.tex
\section{Experimental results}
\label{sec:experiments}

This section presents our experimental results. We evaluate our approach on three tasks: image-to-recipe retrieval (Sec.~\ref{sec:experiments_im2rec}), program generation from recipes and images (Sec.~\ref{sec:experiments_rec2prog}), and image generation from programs (Sec.~\ref{sec:experiments_gan}). 

\mypar{Data.}
We use the Recipe1M\footnote{obtained from http://im2recipe.csail.mit.edu/} dataset~\cite{salvador17cvpr} as standard in previous work
~\cite{fain2019dividing,fu2020mcen,salvador21cvpr,salvador17cvpr,wang2019learning,wang2021cross,zhu19cvpr}. Recipe1M contains 887,706 food images and 1,029,720 cooking recipes split in training (70\%), validation (15\%) and test (15\%) sets. Note that we use only the recipes that have corresponding images (340,831 recipes). For the cooking programs, we use our ground-truth dataset with 3,708 programs.


\begin{figure}[t]
\center
\includegraphics[width=\linewidth]{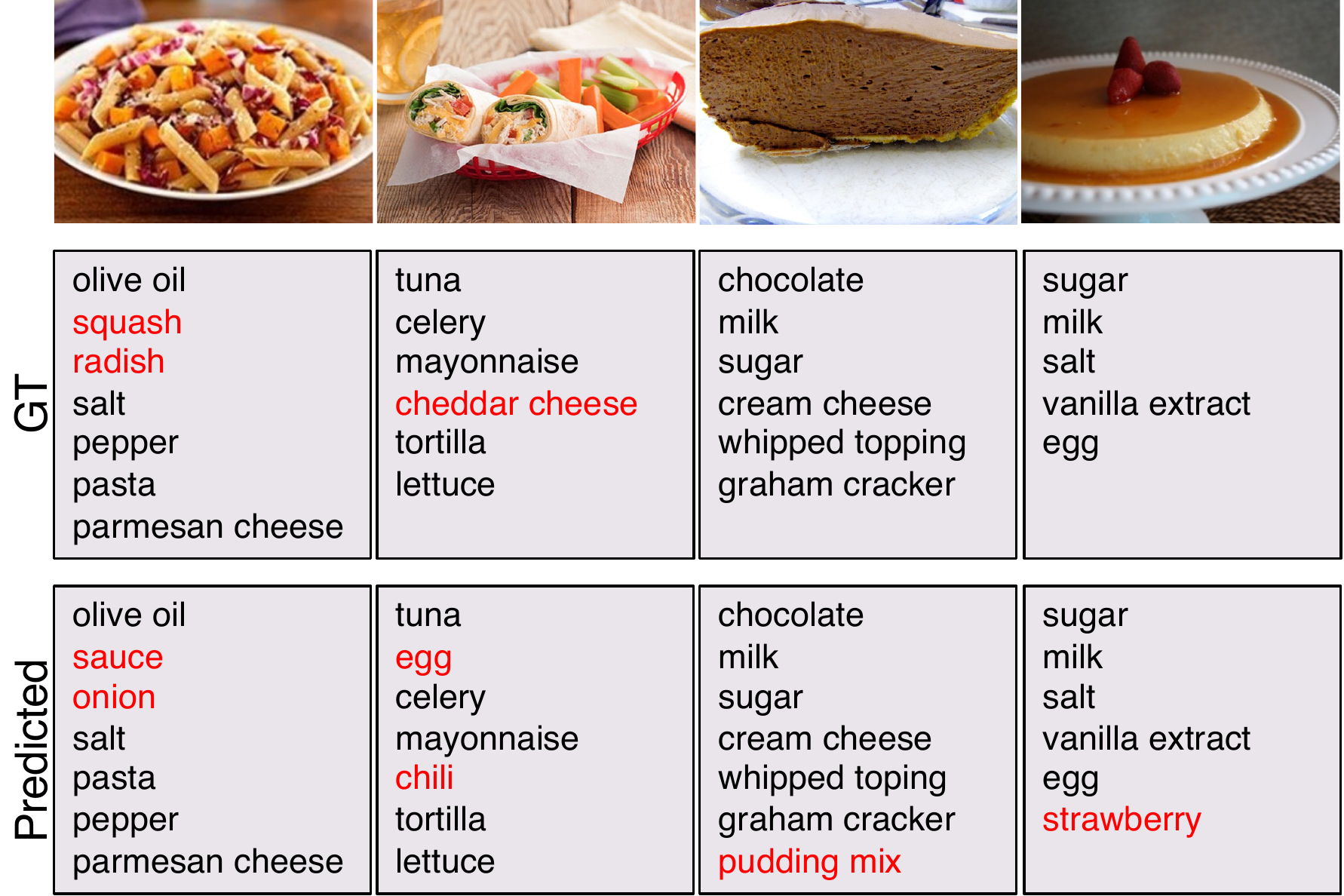}
\caption{\small \textbf{Ingredient prediction from images.} We show qualitative results for the ingredients extracted from generated programs ($P_v$). The ingredients highlighted in red are either FP or FN.}
\label{fig:ingr_classif}
\end{figure}

\mypar{Implementation details.}
Unless stated otherwise, we use the following settings.
${F}_{V}$ is based on ViT-B/16~\cite{dosovitskiy2020image} ($k_v$=12  layers, 12 heads) originally pretrained on ImageNet~\cite{russakovsky15ijcv}.
${F}_{T}$ and $G_P$ are based on~\cite{vaswani17nips}. We use $k_t$=8 encoder layers (8 heads) and 2 decoder layers (4 heads). To handle the imbalance between the image-recipe and the recipe-program pairs, we first pre-train $F_T$ and $G_P$ alone using our annotated recipe-program pairs and obtain pseudo ground-truth programs for the whole Recipe1M.
The images are resized to 256$\times$256 and then cropped to 224$\times$224. During training we perform random crop and horizontal flip augmentation. 
%
We train our model for 50 epochs using the Adam optimizer~\cite{kingma15iclr} with a base learning rate of $10^{-4}$ and a step decay of 0.1 every 20 epochs. We set $\lambda_{ss}=1$, $\lambda_{pv}=0.1$, $\lambda_{pt}=0.1$ and $m=0.3$. All experiments were run on four Nvidia Titan X GPUs.

\subsection{Image-to-recipe retrieval}
\label{sec:experiments_im2rec}

\mypar{Evaluation.}
Following the protocol of Recipe1M~\cite{salvador17cvpr}, we measure the retrieval performance on the test set with median rank (medR) and recall at top K (R@K) for $K={1,5,10}$ on ranking of 1,000 recipe-image pairs. We report the average metrics after repeating experiments 10 times.

\mypar{Recipe components.} We first examine the effect of training and testing the model using different recipe components. In the first three rows of Tab.~\ref{table:retrieval}, we observe that the ingredient list is the most informative component (R@1=39.1). Combining it with the title gives a small boost in performance (+4.5 in R@1), while using the full recipe yields 53.5.

\mypar{Predicting programs.} The first five models were trained without the decoder $G_P$. We observe that training the model using $G_P$ to predict programs from the common embedding space significantly improves the performance (+4.9 R@1).

\mypar{Deeper encoders.} When using deeper encoders (\textit{Base} model with 12 layers), the R@1 accuracy jumps to 66.9.

\mypar{Comparison with state of the art.} In Tab.~\ref{table:retrieval}(bottom), we compare our final model against several approaches. We achieve state-of-the-art performance thanks to the powerful transformer encoders and our program prediction loss. Note that a fair comparison among all methods is hard since they all use different image (ResNet, ResNext, ViT) and text (LSTMs, transformers) encoders. Also, note that previous work~\cite{salvador21cvpr} use multiple transformer encoders for each component of the recipe. Instead, we use only a single text encoder (Fig.~\ref{fig:main_figure}) leading to a much simpler and efficient model. Also, we do not use any unpaired data from Recipe1M unlike most of the existing approaches~\cite{CarvalhoM:SIGIR18,fain2019dividing,salvador21cvpr,salvador17cvpr,wang2019learning,zhu19cvpr}. 

\mypar{ViT features.} Using features from the ``cls token'' as in the original ViT~\cite{dosovitskiy2020image} (instead of average pooling) causes a drop in performance (-2.5 in R@1), as shown in Tab.~\ref{table:ablation} (top).

\mypar{Loss hyperparameters.} In Tab.~\ref{table:ablation}, we perform an ablation study for $\lambda_{pv}$ and $\lambda_{pt}$ showing a small performance drop (-0.6\% with $\lambda_{pv}=1$ and -2.2\% with $\lambda_{pv}=10$ in R@1).

\begin{figure}[t]
\center
\includegraphics[width=\linewidth]{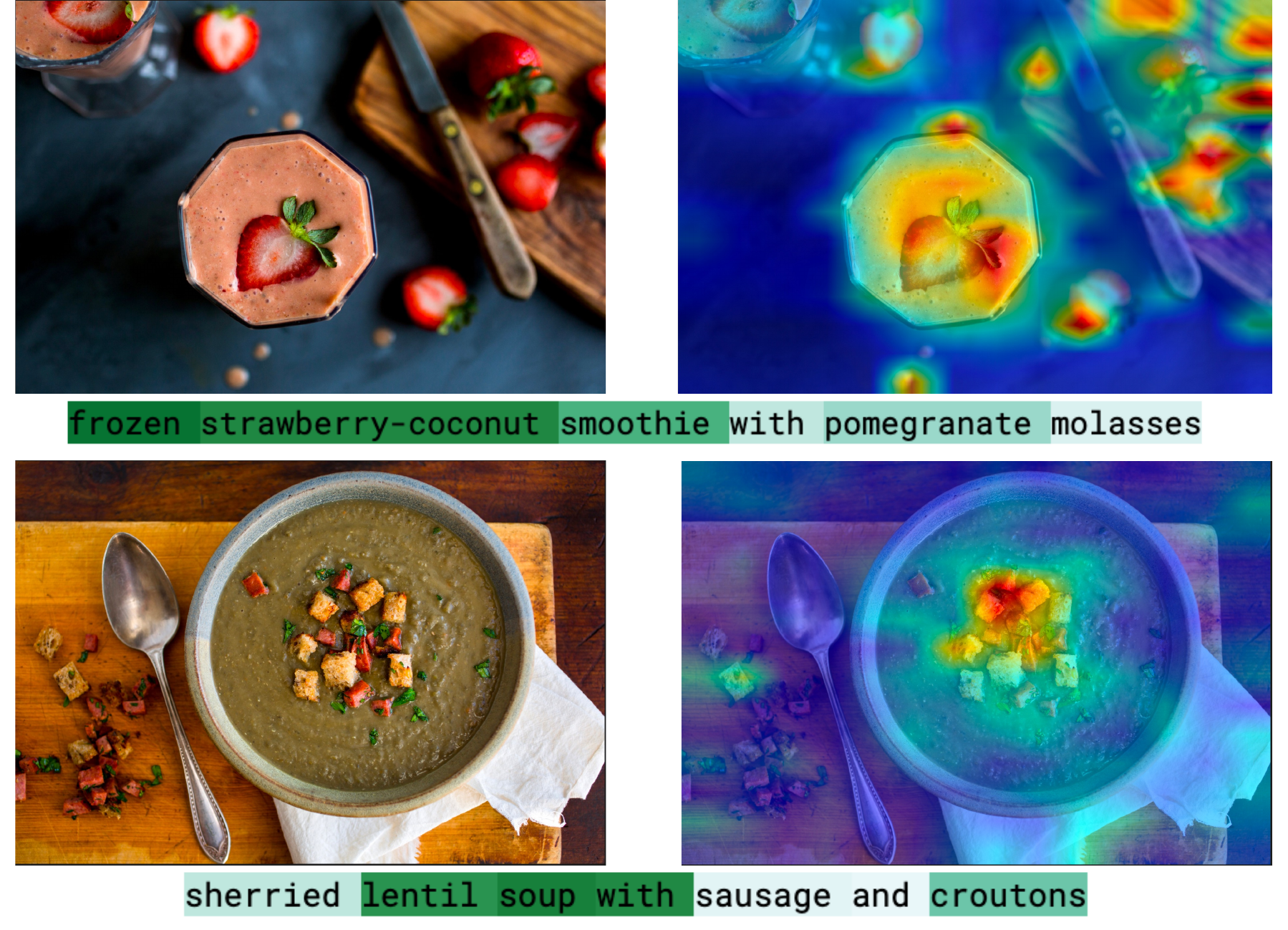}
\caption{\small \textbf{Attention.} We use Attention Rollout~\cite{abnar2020quantifying} to visualize the attention weights on images and on recipe titles.}
\label{fig:attention}
\end{figure}

\begin{figure*}[t]
\center
\includegraphics[width=\linewidth]{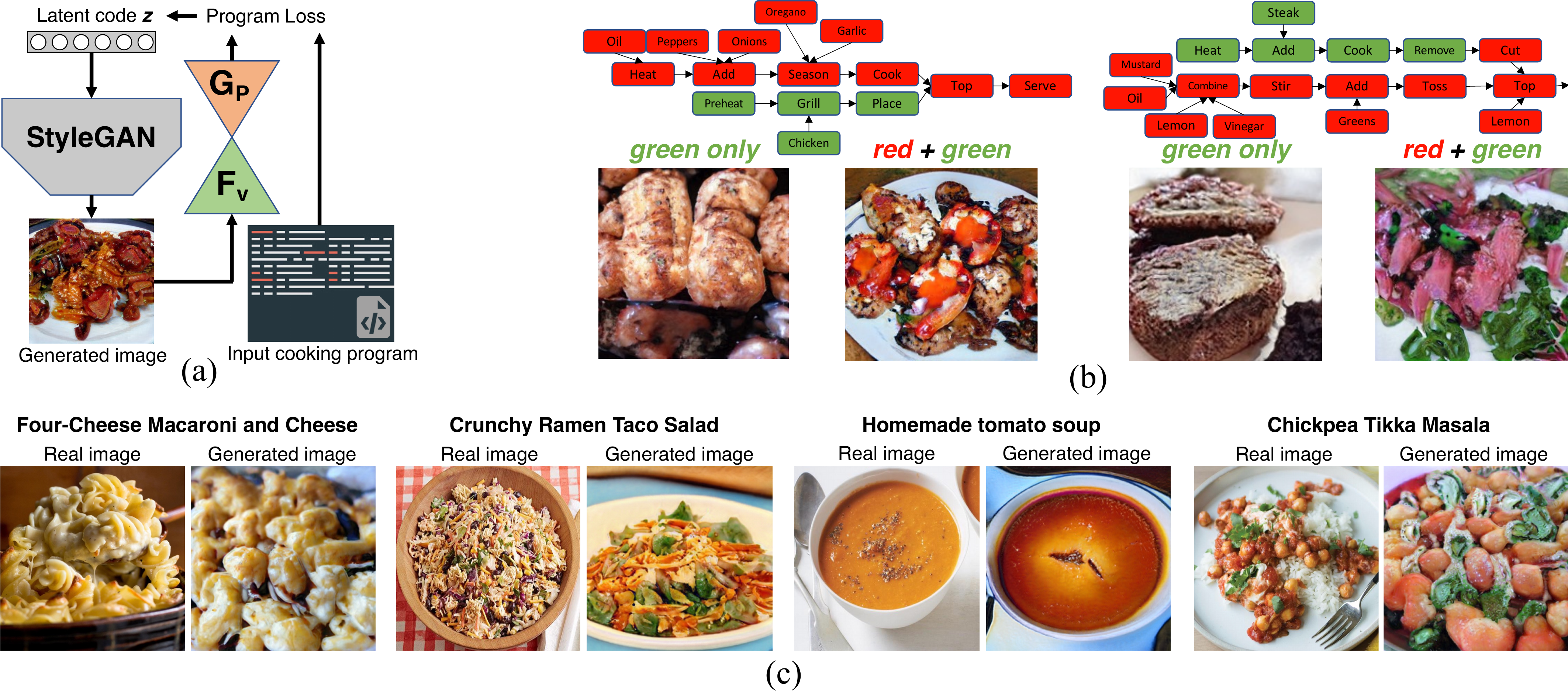}
\caption{\small \textbf{Generating food images from programs.} (a) We optimize the latent code of a GAN by computing a loss between the input program and the predicted one from a generated image. (b) Generating images by manipulating programs. The images on the left of the graphs are generated using only the green nodes, while images on the right are obtained from the full graph. (c) Examples of generated images from ground-truth programs and comparison with the real images.}
\label{fig:gan}
\end{figure*}

\subsection{Program prediction from images and recipes}
\label{sec:experiments_rec2prog}

We evaluate here our model on the task of predicting programs from images or recipes (Fig.~\ref{fig:graph_example}, Tab.~\ref{table:ingredients}).

\mypar{Evaluation.} Instead of considering the program as a sequence and use standard sequence evaluation metrics~\cite{papineni2002bleu}, we turn the programs into graphs and measure the graph edit distance (GED)~\cite{sanfeliu1983distance} wrt to the groundtruth one. GED captures not only the semantic nodes but also their ordering and topology. This evaluation is feasible due to our fixed program vocabulary. Moreover, we extract only the ingredient nodes of the graphs and measure accuracy with the F1 score between the predicted and the ground-truth sets.
We follow the same strategy for the action and tool nodes.

\mypar{Programs from images.} In Tab.~\ref{table:ingredients} (top), we report our results for predicting programs from images and compare with several baselines: 
(a) \textit{Random image}: program from a random image;
(b) \textit{Retrieved image}: program from the top retrieved image using our retrieval system;
(c) \textit{NN (oracle)}: oracle nearest neighbor (NN) program
(e) \textit{Instructions}: our model where we decode the instruction text instead of a program.
Tab.~\ref{table:ingredients} (top) shows that generating programs leads to better results (better ingredient, action and tool prediction) than the strong baselines of decoding sentences or relying on a retrieval system for making predictions.
Interestingly, our model is only slightly below the oracle baseline that simulates an ideal retrieval system.
We also observe that our program loss (minCE) slightly improves results over the standard CE loss. 
Fig.~\ref{fig:ingr_classif} show qualitative results for the ingredient prediction on food
images, while Fig.~\ref{fig:graph_example} shows a predicted program from a recipe and an image.

\mypar{Programs from recipes.} Similarly, in Tab.~\ref{table:ingredients} (bottom), we report our results for predicting programs from recipes and repeat the above baselines. The \textit{text classifier} here is a transformer-based multi-label classifier that predicts ingredients. As expected, predicting programs from recipes is a much simpler task than predicting them from images. Once more, we observe similar trends and our loss yields slightly better results than the standard CE and significantly better than the \textit{Instructions} baseline (+33.9\% F1 for ingredients).



\mypar{Attention.} To further understand how  $F_V$ and $F_T$ process images and text, we visualize their attention weights following~\cite{abnar2020quantifying}. In Fig.~\ref{fig:attention} we observe that both encoders focus on semantically similar concepts across the two domains.

\subsection{Image generation}
\label{sec:experiments_gan}

We show an interesting application of image generation conditioned on cooking programs using our model. We first train a StyleGANv2~\cite{karras2020analyzing} on the images of the Recipe1M dataset~\cite{salvador17cvpr}. Given an initial latent vector $z_0$, we use the GAN to generate an image. The image goes through the encoder $F_V$ and the decoder $G_P$ to obtain a program. We compute the loss between this program and the desired input one. We use the loss to optimize $z$ via backpropagation. Similar approaches have been proposed using the recently introduced CLIP~\cite{radford2021learning} model to drive image generation using text sentences~\cite{bau2021paint,patashnik2021styleclip}. Fig.~\ref{fig:gan} (left) illustrates our approach while in Fig.~\ref{fig:gan} (right) we show examples of generated images using a full program or a part of it (green nodes only). We observe that the generated images are realistic and capture plausibly the content of the input program.

%% file: sec6_conc.tex
\section{Conclusions}
\label{sec:concl}

We proposed to model cooking recipes with cooking programs.
We designed a program scheme and annotated a set of programs for cooking recipes and we presented an approach for learning to predict programs from food images and recipes. Experimental results showed that projecting the common space between images and recipes into programs can improve retrieval results.
Finally, we showed how we can generate food images by manipulating a program.
%
%
We hope that programs will open new directions such as allowing agents to execute recipes or allowing us to extract common-sense knowledge for food from graphs.

\mypar{Limitations and societal impact.}
In this work, we do not go beyond predicting the cooking procedure via programs. However, predicting nutritional value, estimating calories and their impact in our health is an important topic. Moreover, inaccurate predicted programs might not be able to be executed or lead to inedible food. Also, ingredient prediction models should be applied consciously especially in user cases with food allergies.
Future work involves an analysis on the potential biases that the programs or our training data might have (e.g. towards unhealthy food dishes, towards western world with underrepresented cuisines) and the impact that this might have on the food industry. 

\mypar{Acknowledgments.} This work is supported by Nestl\'e.